\documentclass{article}
\usepackage{microtype}
\usepackage{multirow}
\usepackage{float}
\usepackage{wrapfig}
\usepackage{graphicx}
\usepackage{caption}
\usepackage{subcaption}
\usepackage{booktabs}
\usepackage{hyperref}

\usepackage{colortbl}
\usepackage{algorithm}
\usepackage[noend]{algorithmic}
\usepackage{enumitem}
\usepackage{bbding}
\usepackage{fancyvrb}
\usepackage[scaled=0.85]{DejaVuSansMono}
\usepackage{listings}
\definecolor{codegreen}{rgb}{0,0.5,0}
\definecolor{codered}{rgb}{0.7,0.1,0.1}
\definecolor{codegray}{rgb}{0.5,0.5,0.5}
\definecolor{codepurple}{rgb}{0.58,0,0.82}
\definecolor{backcolour}{rgb}{1,1,1}
\definecolor{gred}{rgb}{0.859,0.267,0.216}
\definecolor{ggreen}{rgb}{0.059,0.616,0.345}
\definecolor{gblue}{rgb}{0.259,0.522,0.957}
\definecolor{gyellow}{rgb}{0.957,0.706,0}
\definecolor{gpurple}{rgb}{0.565,0.173,0.894}
\lstdefinestyle{python}{
    language=Python,
    backgroundcolor=\color{backcolour},   
    commentstyle=\color{codered}\textit,
    keywordstyle=\bfseries\color{codegreen},
    numberstyle=\tiny\color{codegray},
    stringstyle=\color{codepurple},
    basicstyle=\ttfamily\footnotesize,
    breakatwhitespace=false,         
    breaklines=true,                 
    captionpos=b,                    
    keepspaces=true,                 
    numbers=left,                    
    numbersep=5pt,                  
    showspaces=false,                
    showstringspaces=false,
    showtabs=false,                  
    tabsize=2,
    fancyvrb=true
}
\newenvironment{codesnippet}
  { \VerbatimEnvironment%
    \begin{Verbatim} }
  { \end{Verbatim}  }
\definecolor{darkgreen}{rgb}{0.09, 0.45, 0.27}
\usepackage[accepted]{icml2023}
\usepackage{ragged2e}
\usepackage{enumerate}
\usepackage{enumitem}
\usepackage{colortbl}
\usepackage{amsmath}
\usepackage{amssymb}
\usepackage{mathtools}
\usepackage{amsthm}
\usepackage{changepage}

\theoremstyle{plain}

\theoremstyle{definition}

\theoremstyle{remark}

\icmltitlerunning{Revisiting a Learning-from-Scratch Baseline}

\begin{document}

\twocolumn[
\icmltitle{On Pre-Training for Visuo-Motor Control:\\Revisiting a Learning-from-Scratch Baseline}

\icmlsetsymbol{equal}{*}
\icmlsetsymbol{equaladvising}{$\dagger$}

\begin{icmlauthorlist}
\icmlauthor{Nicklas Hansen}{ucsd,equal}
\icmlauthor{Zhecheng Yuan}{tsinghua,equal}
\icmlauthor{Yanjie Ze}{ucsd,sjtu,equal}
\icmlauthor{Tongzhou Mu}{ucsd,equal} \\
\vspace*{5pt}
\icmlauthor{Aravind Rajeswaran}{meta,equaladvising}
\icmlauthor{Hao Su}{ucsd,equaladvising}
\icmlauthor{Huazhe Xu}{tsinghua,sqzi,equaladvising}
\icmlauthor{Xiaolong Wang}{ucsd,equaladvising}
\end{icmlauthorlist}

\icmlaffiliation{ucsd}{University of California San Diego}
\icmlaffiliation{meta}{Meta AI}
\icmlaffiliation{tsinghua}{Tsinghua University}
\icmlaffiliation{sjtu}{Shanghai Jiao Tong University}
\icmlaffiliation{sqzi}{Shanghai Qi Zhi Institute}

\icmlcorrespondingauthor{Nicklas Hansen}{nihansen@ucsd.edu}
\icmlkeywords{Reinforcement Learning, Visuo-Motor Control, Pre-Training}

\vskip 0.3in
]

\renewcommand{\icmlEqualContribution}{\textsuperscript{*}Equal contribution \textsuperscript{$\dagger$}Equal advising}
\printAffiliationsAndNotice{\icmlEqualContribution}

\begin{abstract}
In this paper, we examine the effectiveness of pre-training for visuo-motor control tasks. We revisit a simple Learning-from-Scratch (LfS) baseline that incorporates data augmentation and a shallow ConvNet, and find that this baseline is surprisingly competitive with recent approaches (PVR, MVP, R3M) that leverage frozen visual representations trained on large-scale vision datasets -- across a variety of algorithms, task domains, and metrics in simulation and on a real robot. Our results demonstrate that these methods are hindered by a significant domain gap between the pre-training datasets and current benchmarks for visuo-motor control, which is  alleviated by finetuning. Based on our findings, we provide recommendations for future research in pre-training for control and hope that our simple yet strong baseline will aid in accurately benchmarking progress in this area.\footnote{Code: {\tiny \url{https://github.com/gemcollector/learning-from-scratch}}.}
\end{abstract}

\section{Introduction}
\label{sec:introduction}
Large-scale pre-training has delivered promising results in computer vision~\citep{Doersch2015UnsupervisedVR, He2020MomentumCF, Oord2018RepresentationLW, alayrac2022flamingo} and natural language processing~\citep{Devlin2019BERTPO, Brown2020LanguageMA, radford2021learning, chowdhery2022palm}. Recent works have extended the pre-training paradigm to visuo-motor control by leveraging pre-trained visual representations for policy learning~\citep{Parisi2022TheUE, Nair2022R3MAU, Xiao2022MaskedVP,Ze2022rl3d,yuan2022pretrained}. These works train a visual representation using large out-of-domain vision datasets like ImageNet~\cite{Russakovsky2015ImageNetLS} and Ego4D~\citep{grauman2022ego4d}, and freeze the vision model weights for downstream policy learning. When compared to simple Learning-from-Scratch~(LfS) methods for visuo-motor control, these works find that frozen pre-trained representations help achieve higher sample efficiency and/or asymptotic performance across various task domains.

\begin{figure*}[h]
    \centering
    \includegraphics[width=0.945\linewidth]{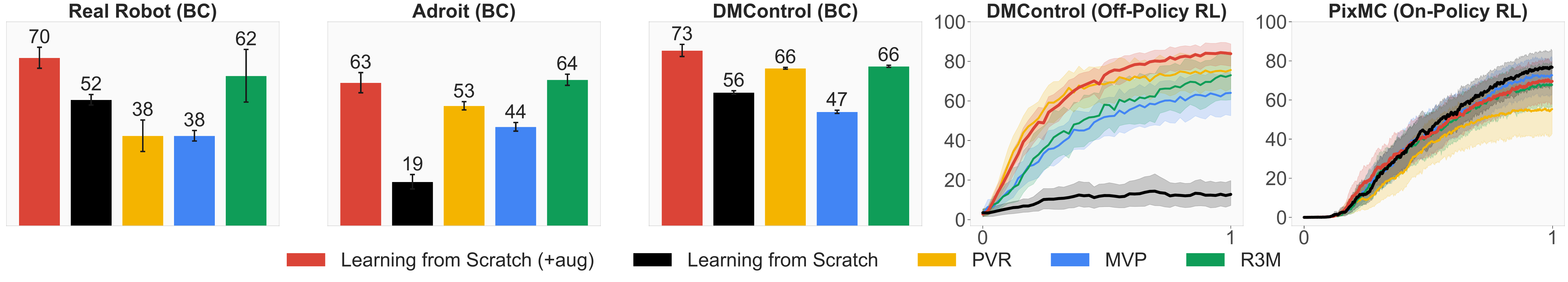}
    \vspace{-0.175in}
    \caption{\textbf{Pre-training vs. Learning-from-Scratch (LfS).} Success rate (real robot, Adroit, PixMC) and normalized return (DMControl) in each of the four task domains that we consider (aggregated across tasks). BC simulation results are averages of top-3 evaluations over 100 epochs \citep{Parisi2022TheUE}, and RL results are reported as a function of environment steps \citep{yarats2021mastering, Xiao2022MaskedVP}, normalized to the interval $(0,1)$ as total steps differ between tasks. We evaluate strong yet simple LfS baselines \citep{yarats2021mastering, Hansen2021StabilizingDQ} and find them to be competitive with recent frozen pre-trained representations. Mean and $95\%$ CIs over 5 seeds.}
    \label{fig:main-result}
    \vspace{-0.0275in}
\end{figure*}

\begin{figure*}[t]
    \centering
    \begin{minipage}{0.2455\textwidth}
        \centering
        \includegraphics[width=0.49\textwidth]{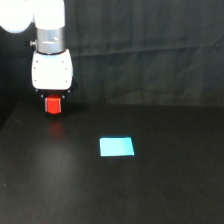}
        \includegraphics[width=0.49\textwidth]{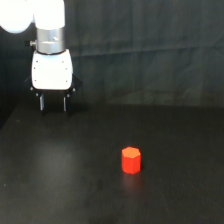}\\ \vspace{-0.025in}
        {\small \textbf{Real robot}}
    \end{minipage}
    \begin{minipage}{0.2455\textwidth}
        \centering
        \includegraphics[width=0.49\textwidth]{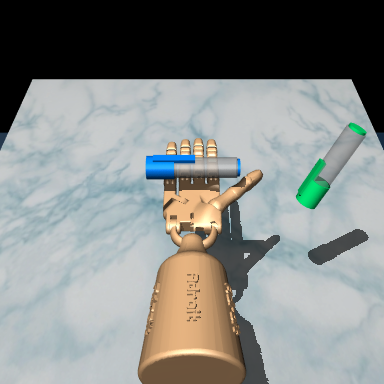}
        \includegraphics[width=0.49\textwidth]{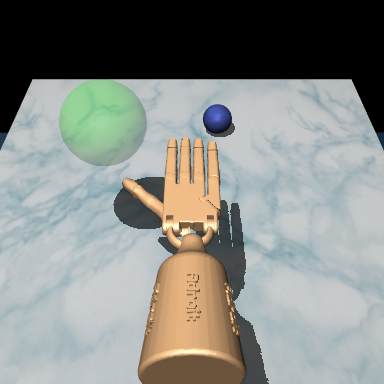}\\ \vspace{-0.025in}
        {\small \textbf{Adroit}}
    \end{minipage}
    \begin{minipage}{0.2455\textwidth}
        \centering
        \includegraphics[width=0.49\textwidth]{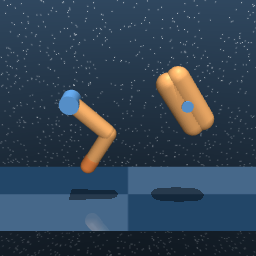}
        \includegraphics[width=0.49\textwidth]{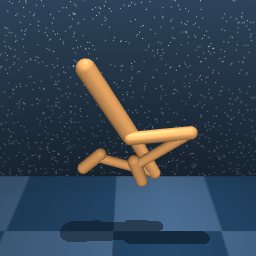}\\ \vspace{-0.025in}
        {\small \textbf{DMControl}}
    \end{minipage}
    \begin{minipage}{0.2455\textwidth}
        \centering
        \includegraphics[width=0.49\textwidth]{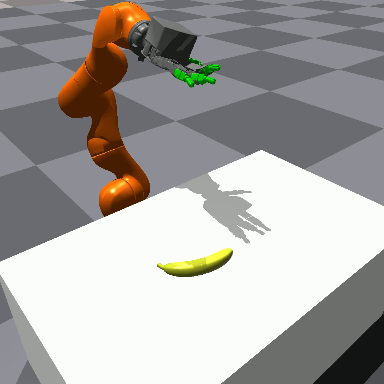}
        \includegraphics[width=0.49\textwidth]{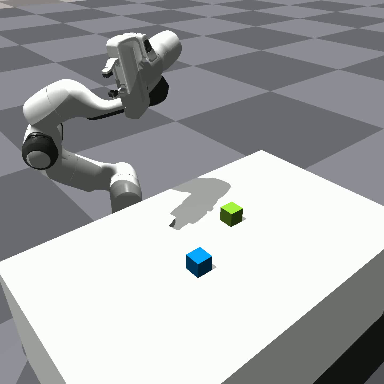}\\ \vspace{-0.025in}
        {\small \textbf{PixMC}}
    \end{minipage}
    \vspace{-0.125in}
    \caption{\textbf{Tasks.} We consider challenging and diverse visuo-motor control tasks spanning $\mathbf{4}$ domains, from left to right: a real robot setup (manipulation), Adroit (dexterous manipulation), DMControl (locomotion, manipulation), and PixMC (manipulation). Our experimental setups in simulation are adopted from PVR, MVP, and R3M, and our real setup is similar to that of R3M. We consider a total of 17 tasks.}
    \label{fig:tasks}
    \vspace{-0.135in}
\end{figure*}

However, there exists a rich body of work on strategies to improve performance of LfS methods, such as auxiliary self-supervised representation learning~\cite{Srinivas2020CURLCU, Schwarzer2021DataEfficientRL} or using carefully curated data augmentations~\cite{Laskin2020ReinforcementLW, Kostrikov2021ImageAI, yarats2021mastering, raileanu2020automatic, Hansen2021StabilizingDQ}. To gain a sharp understanding of the advantages of visual pre-training for visuo-motor control, it is necessary to establish strong LfS baselines.

Towards this end, we adopt the experimental setups of prior works without modification, and implement strong LfS baselines that leverage shallow ConvNet encoders and random shift data augmentation \citep{Kostrikov2021ImageAI, yarats2021mastering}. Surprisingly, we find that this modified LfS baseline can achieve results competitive with prior works that leverage frozen pre-trained visual representations. While our contributions are incremental in nature, we believe that our work contains must-know insights for anyone working on pre-trained representations for visuo-motor control.

We evaluate our approach across a variety of task domains, algorithm classes, and evaluation metrics. Specifically, we examine $\mathbf{4}$ task domains (Adroit~\citep{Rajeswaran-RSS-18}, DMControl~\citep{deepmindcontrolsuite2018}, PixMC~\citep{Xiao2022MaskedVP}, and a real robot setup), $\mathbf{3}$ algorithm classes: imitation learning~(behavior cloning), on-policy RL~(PPO~\citep{Schulman2017ProximalPO}), and off-policy RL~(DrQ-v2~\citep{yarats2021mastering}), and multiple evaluation metrics including sample-efficiency, asymptotic performance, visual robustness, and computational cost. To our surprise, our carefully designed LfS baseline is found to be competitive with frozen pre-trained representations across most settings and metrics, and in some cases even outperforms them. At present, frozen pre-trained representations are found to mostly be advantageous in terms of computational cost.

We remain optimistic that \emph{pre-trained representations will play an important and increasingly larger role} in visuo-motor control as the paradigm matures. At the same time, we believe that setting a simple yet strong baseline will help accurately benchmark progress in this area. Based on our empirical findings, we provide recommendations for future research in pre-training for control. In particular, we conjecture that current benchmark tasks are not well suited to reap the benefits of pre-trained representations, since they do not require any visual generalization. As the community builds better benchmarks and harder tasks that require both visual and policy generalization, we conjecture that pre-trained representations will play an increasingly important role. Additionally, our results indicate that current pre-trained representations suffer from a substantial domain gap by pre-training on large-scale real-world data and benchmarking on predominantly simulated environments, which we find can be alleviated with careful in-domain finetuning based on our LfS insights. In the following sections, we detail each method, experimental setup, and results, and conclude with a broader discussion on the implications of our findings.

\section{Methods}
\label{sec:methods}
Comparing two \emph{paradigms} fairly is difficult, and comparing LfS with pre-trained representations is no exception.
To help narrow our scope, we focus on \emph{representative methods} from each paradigm: a simple Learning-from-Scratch (LfS) method that uses a shallow ConvNet and data augmentation, as well as three \textbf{frozen} visual representations trained on large-scale out-of-domain vision datasets (PVR \citep{Parisi2022TheUE}, MVP \citep{Xiao2022MaskedVP}, R3M \citep{Nair2022R3MAU}). These three visual representations were proposed concurrently and represent the present state-of-the-art in pre-training for visuo-motor control. \emph{We choose to freeze the pre-trained representations to be consistent with their original formulations}. The three pre-trained representations that we consider have been shown to outperform common representations such as supervised learning and MoCo-v2~\citep{He2020MomentumCF} pre-training on ImageNet \citep{Russakovsky2015ImageNetLS}. In the following, we provide a more detailed description of each pre-trained representation, as well as our proposed LfS baseline. See Table \ref{tab:overview-methods} for an overview of the three pre-trained representations that we consider.

\begin{table}[t]
    \centering
    \vspace*{-7pt}
    \caption{\textbf{Overview of frozen pre-trained representations.} We summarize key design choices for each of the three pre-trained representations proposed in prior work, as well as which algorithm they considered in downstream tasks. \emph{Jitter} denotes whether color jitter augmentation was applied during pre-training; this detail pertains to our visual robustness experiments in Section \ref{sec:experiments-results}.}
    \vspace{0.025in}
    \label{tab:overview-methods}
    \resizebox{0.475\textwidth}{!}{%
    \begin{tabular}[b]{lcccc|c}
    & \multicolumn{4}{c}{Pre-training} & Policy \\
    \toprule
    Method & Repr. & Encoder & Dataset & Jitter & Algo.
    \\ \midrule
    \textcolor{gyellow}{$\bullet$} PVR & MoCo-v2 & ResNet50 & ImageNet & \textcolor{ggreen}{\Checkmark} & BC \\
    \textcolor{gblue}{$\bullet$} MVP & MAE & ViT-S & HOI & \textcolor{gred}{\XSolidBrush} & PPO \\
    \textcolor{ggreen}{$\bullet$} R3M & Multi-loss & ResNet50 & Ego4D & \textcolor{gred}{\XSolidBrush} & BC \\ \bottomrule
    \end{tabular}
    }
    \vspace{-0.175in}
\end{table}

\begin{figure*}[t]
    \centering
    \begin{minipage}{0.495\textwidth}
        \centering
        \includegraphics[width=0.235\textwidth]{Figures/viz_real_robot/reach/0.png}
        \includegraphics[width=0.235\textwidth]{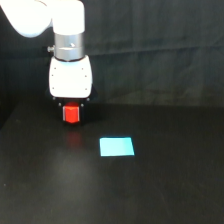}
        \includegraphics[width=0.235\textwidth]{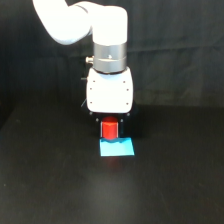}
        \includegraphics[width=0.235\textwidth]{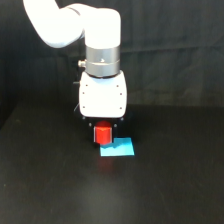}\\ \vspace{-0.025in}
        {\small \textbf{Reach}}
    \end{minipage}
    \begin{minipage}{0.495\textwidth}
        \centering
        \includegraphics[width=0.2425\textwidth]{Figures/viz_real_robot/pick/0.png}
        \includegraphics[width=0.2425\textwidth]{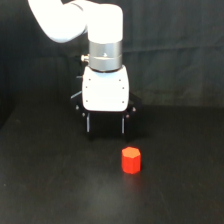}
        \includegraphics[width=0.2425\textwidth]{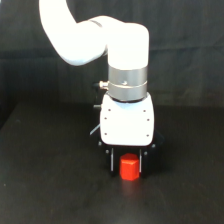}
        \includegraphics[width=0.2425\textwidth]{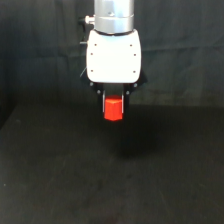}\\ \vspace{-0.025in}
        {\small \textbf{Pick}}
    \end{minipage}
    \vspace{-0.075in}
    \caption{\textbf{Real robot tasks.} Sample trajectories for each of the two real robot tasks that we consider, \emph{reach} and \emph{pick}. Visualizations correspond to raw RGB observations at key frames. Actual episode length is 50. Trajectories are generated using \textcolor{gred}{$\bullet$} LfS (+aug) with 10 and 20 demonstrations, respectively, collected by a human teleoperator. We evaluate methods on 20 trials per task and across 2 random seeds.}
    \label{fig:real-robot-tasks}
    \vspace{-0.05in}
\end{figure*}

\textcolor{gyellow}{$\bullet$} \textbf{PVR} investigates the efficacy of frozen pre-trained representations for Behavior Cloning (BC) in a variety of control tasks, and proposes a variant of Momentum Contrast (MoCo-v2; \citep{He2020MomentumCF}) that fuses features from multiple layers. Specifically, the PVR model is a MoCo-v2 representation with a ResNet50 \citep{He2015DeepRL} backbone trained on ImageNet \citep{Russakovsky2015ImageNetLS}, with intermediate layers fused together with final output features via a second finetuning stage. Combining features from early and later layers of the network encourages the PVR model to retain spatial granularity as well as scene-level semantic information. The PVR representation is trained on individual frames without leveraging temporal-sequential information. During pre-training, PVR applies random crop, horizontal flip, gray-scale, and color jitter augmentations. We use the publicly available PVR model in our experiments.

\textcolor{gblue}{$\bullet$} \textbf{MVP} concurrently studies the efficacy of frozen pre-trained representations for on-policy RL using Proximal Policy Optimization (PPO; \citep{Schulman2017ProximalPO}), and propose to train a Masked Autoencoder \citep{He2022MaskedAA} visual representation on individual frames from a large (700K frames) human interaction dataset (referred to as \emph{HOI}) sourced from multiple existing datasets. Concretely, the MVP model uses a Vision Transformer (ViT; \citep{Dosovitskiy2021AnII}) backbone that partitions frames into $16\times16$ patches. The MVP representation is trained on individual frames without leveraging temporal-sequential information. The parameter count of the MVP encoder (ViT-S; $22$M) is comparable to that of PVR and R3M. During pre-training, MVP applies random crop and horizontal flip augmentations. We use the publicly available MVP model in our experiments\footnote[2]{We acknowledge that a newer set of MVP models have been released concurrently with our work. While improved downstream task performance can be anticipated for the new models, we expect our main conclusions to remain unchanged.}.

\textcolor{ggreen}{$\bullet$} \textbf{R3M} proposes to pre-train a ResNet50 backbone using a combination of time-contrastive learning \citep{Sermanet2016UnsupervisedPR}, video-language alignment, and L1 regularization that encourages sparse and compact representations, on 3,500 hours of human interaction video data from the Ego4D dataset \citep{grauman2022ego4d}. In contrast to PVR and MVP, R3M \emph{does} leverage the temporal-sequential nature of video data. During pre-training, R3M only applies random crop augmentations. We use the publicly available R3M model in our experiments.

We evaluate these three pre-trained representations -- PVR, MVP, and R3M -- against our simple yet strong LfS baseline that uses data augmentation and a shallow ConvNet encoder. To demonstrate the importance of data augmentation in representation learning for control, we include two LfS baselines -- with and without use of data augmentation -- which we describe in the following.

\textcolor{black}{$\bullet$} \textbf{LfS} \emph{(no aug)} uses a shallow ConvNet encoder that consists of 4-6 layers (depending on the experimental setup in which it is applied) of 2D-convolutions with ReLU activations. Each of our three encoder implementations are adopted from prior work and are widely accepted by the RL community. Specifically, the LfS encoder in our BC experiments is identical to that of PVR \citep{Parisi2022TheUE}, the LfS encoder that we use for off-policy RL is equivalent to that of \citet{yarats2019improving, Srinivas2020CURLCU, Laskin2020ReinforcementLW, Kostrikov2021ImageAI, yarats2021mastering}, and our on-policy RL LfS baseline is identical to that of \citet{hansen2022modem}.

\textcolor{gred}{$\bullet$} \textbf{LfS} \emph{(+aug)} uses an architecture identical to that of \textcolor{black}{$\bullet$}LfS. However, it is well documented in literature on visual RL that use of data augmentation is critical to the performance and visual robustness of LfS \citep{Laskin2020ReinforcementLW, Kostrikov2021ImageAI, yarats2021mastering, hansen2021softda, raileanu2020automatic, Hansen2021StabilizingDQ, yuan2022pretrained}. To accurately reflect progress in LfS approaches, our main point of comparison is a LfS method that uses random shift augmentation \citep{Kostrikov2021ImageAI} in addition to its shallow ConvNet encoder, which has demonstrated strong empirical performance on a variety of task domains. As our experiments reveal, use of data augmentation is also surprisingly effective for learning behavior cloning policies, although it is not commonly used in this setting.

The reader is referred to our respective experimental setups in Section \ref{sec:experiments-setup} for a per-algorithm description of our proposed LfS baselines.

\section{Experimental setup}
\label{sec:experiments-setup}
We propose a set of strong LfS baselines that span $\mathbf{3}$ classes of algorithms: imitation learning (behavior cloning), on-policy RL (PPO \citep{Schulman2017ProximalPO}), and off-policy RL (DrQ-v2 \citep{yarats2021mastering}), and consider a total of $\mathbf{17}$ tasks across $\mathbf{4}$ domains: Adroit \citep{Rajeswaran-RSS-18} (dexterous manipulation; 2 tasks $\times$ 2 views), DMControl \citep{deepmindcontrolsuite2018} (locomotion and control; 5 tasks), PixMC \citep{Xiao2022MaskedVP} (robotic manipulation; 8 tasks), and a real robot setup (robotic manipulation; 2 tasks). Figure \ref{fig:tasks} shows sample tasks from each domain; see Appendix \ref{sec:appendix-tasks} for a detailed description of all tasks. Sample trajectories for each of the two real robot tasks are shown in Figure \ref{fig:real-robot-tasks}. Importantly, we do \emph{not} propose a new benchmark for pre-trained representations, but rather base our experiments on the public implementations of PVR, MVP, and DrQ-v2, and \emph{meticulously follow their respective experimental setups}. We make no changes to hyperparameters. This strict experimental setup ensures that pre-trained representations are evaluated in favorable conditions (for which they were originally proposed). Our code is available at {\small\url{https://github.com/gemcollector/learning-from-scratch}}. We provide the full details of our experimental setup in the appendices, and summarize it as follows:

$\boldsymbol{-}$ \textbf{Behavior Cloning (BC).} We consider two simulation domains -- Adroit and DMControl -- used in PVR, in addition to our real robot setup. Observations are $256\times256$ RGB images (center-cropped to $224\times224$) with no access to proprioceptive information. In simulation, policies are trained with BC on 100 demonstrations per task; we use the exact demonstration dataset that PVR used\footnote[3]{While the demonstration dataset used in PVR is not publicly available, the authors kindly provided us with the demonstrations in response to our private inquiry. We thank the authors for that.}, \emph{i.e.}, Adroit demonstrations are generated by oracle (state-based) DAPG \citep{Rajeswaran-RSS-18} policies, and DMControl demonstrations are generated by oracle DDPG \citep{Lillicrap2016ContinuousCW} policies. We use 10-20 demonstrations in the real world depending on the task, but otherwise follow the same experimental setup as in simulation. The original LfS baseline in PVR uses a shallow ConvNet encoder; we refer to this baseline simply as $\bullet$\emph{LfS}. {Our improved LfS baseline additionally uses random shift augmentation} \citep{Kostrikov2021ImageAI, yarats2021mastering} during learning, and we refer to this baseline as \textcolor{gred}{$\bullet$}\emph{LfS (+aug)}. Data augmentation is relatively underexplored in BC literature, but we find that it works surprisingly well. In addition to PVR, we also compare with frozen MVP and R3M representations. Consistent with the experimental setup in PVR, we measure the policy performance with success rate in the case of Adroit (and our real setup), and episode return in DMControl. Policies are evaluated every two epochs for a total of 100 epochs in simulation, and we report the average performance over the 3 best epochs over the course of learning. We find that 1 epoch is sufficient for our real robot experiments, where we evaluate for 20 trials per method per task, across 2 random seeds.\vspace{0.05in}\\
$\boldsymbol{-}$ \textbf{On-policy RL.} We reproduce the results of MVP on their proposed PixMC robotic manipulation benchmark. Observations are $224\times224$ RGB images and also include proprioceptive information. The original LfS baseline uses a small ViT \citep{Dosovitskiy2021AnII} encoder. We propose \emph{two} improved LfS baselines for this setting: \emph{(1)} {an LfS baseline that uses a shallow ConvNet encoder and \emph{no} data augmentation}, referred to as $\bullet$\emph{LfS}, and \emph{(2)} {an LfS baseline that additionally applies random shift augmentation} in critic learning, referred to as \textcolor{gred}{$\bullet$}\emph{LfS (+aug)}. Following prior work \citep{Hansen2021StabilizingDQ, raileanu2020automatic}, we do not augment value targets. In addition to (frozen) MVP, we also compare with frozen PVR and R3M representations. Following the setup in MVP, we use the success rate of the policy as the metric for comparison.\vspace{0.05in}\\
$\boldsymbol{-}$ \textbf{Off-policy RL.} We reproduce the results of the state-of-the-art LfS method DrQ-v2 on the same DMControl tasks as used in PVR. Observations are $84\times84$ RGB images with no access to proprioceptive information; we upsample observations to $224\times224$ when using pre-trained representations. {DrQ-v2 uses a shallow ConvNet encoder and random shift augmentation by default}, and we refer to this baseline as \textcolor{gred}{$\bullet$}\emph{LfS (+aug)}. We compare DrQ-v2 to two alternatives: \emph{(1)} not using data augmentation (simply denoted $\bullet$\emph{LfS}), and \emph{(2)} removing data augmentation \textbf{and} additionally replacing the LfS encoder with a frozen pre-trained representation, denoted by their representation names (PVR, R3M, MVP) respectively. Following prior work on DMControl, we use (normalized) return as the metric for comparison.

\begin{table}[t]
    \centering
    \vspace{-0.075in}
    \caption{\textbf{Behavior Cloning: LfS vs. frozen pre-trained visual representations.} Success rate (Adroit, real robot) and unnormalized return (DMControl) of LfS and the \textbf{best} result obtained with a pre-trained representation, \emph{i.e.}, for each task we report $\max\{\textrm{PVR}, \textrm{MVP}, \textrm{R3M}\}$. A well-designed LfS method is competitive with frozen pre-trained representations across all tasks.}
    \label{tab:il}
    \vspace{0.075in}
    \resizebox{0.455\textwidth}{!}{%
    \begin{tabular}[b]{lcccc}
        \toprule
        \textbf{~~~~~~~~Method} & $\bullet$ LfS & \textcolor{gred}{$\bullet$} LfS & \textbf{Best} \\
        \textbf{Task} & (\emph{no aug}) & (\emph{+aug}) & pre-training
        \\ \midrule
        \textcolor{darkgray}{\underline{Adroit}} & & & \\
        Pen         & $22.0\scriptstyle{\pm 4.0}$  & $74.8\scriptstyle{\pm 5.0}$   & $\mathbf{81.3\scriptstyle{\pm 4.0}}$ \\
        Relocate    & $16.9\scriptstyle{\pm 3.5}$   & $\mathbf{51.4\scriptstyle{\pm 7.7}}$   & $47.5\scriptstyle{\pm 2.6}$ \\ \midrule
        \textcolor{darkgray}{\underline{DMControl}} & & & \\
        Finger Spin & $647.6\scriptstyle{\pm 6.9}$   & $661.4\scriptstyle{\pm 22.6}$   & $\mathbf{698.5\scriptstyle{\pm 8.4}}$ \\
        Reacher Easy& $261.3\scriptstyle{\pm 27.6}$ & $\mathbf{657.4\scriptstyle{\pm 44.3}}$   & $615\scriptstyle{\pm 27.0}$ \\
        Cheetah Run & $469.8\scriptstyle{\pm 30.0}$   & $448.9\scriptstyle{\pm 56.4}$   & $\mathbf{557.6\scriptstyle{\pm 18.4}}$ \\
        Walker Stand& $699.0\scriptstyle{\pm 65.0}$ & $\mathbf{875.5\scriptstyle{\pm 20.4}}$   & $818.2\scriptstyle{\pm 19.4}$ \\
        Walker Walk & $699.4\scriptstyle{\pm 15.2}$   & $\mathbf{791.6\scriptstyle{\pm 17.8}}$ & ${788.0\scriptstyle{\pm 10.2}}$ \\ \midrule
        \textcolor{darkgray}{\underline{Real robot}} & & & \\
        Reach         & $80.0\scriptstyle{\pm 0.0}$  & $85.0\scriptstyle{\pm 5.0}$   & $\mathbf{90.0\scriptstyle{\pm 10.0}}$ \\
        Pick          & $25.0\scriptstyle{\pm 5.0}$  & $\mathbf{55.0\scriptstyle{\pm 5.0}}$   & $35.0\scriptstyle{\pm 15.0}$ \\ \bottomrule
    \end{tabular}
    }
    \vspace{-0.05in}
\end{table}
\begin{figure*}[t]
    \centering
    \includegraphics[width=0.935\linewidth]{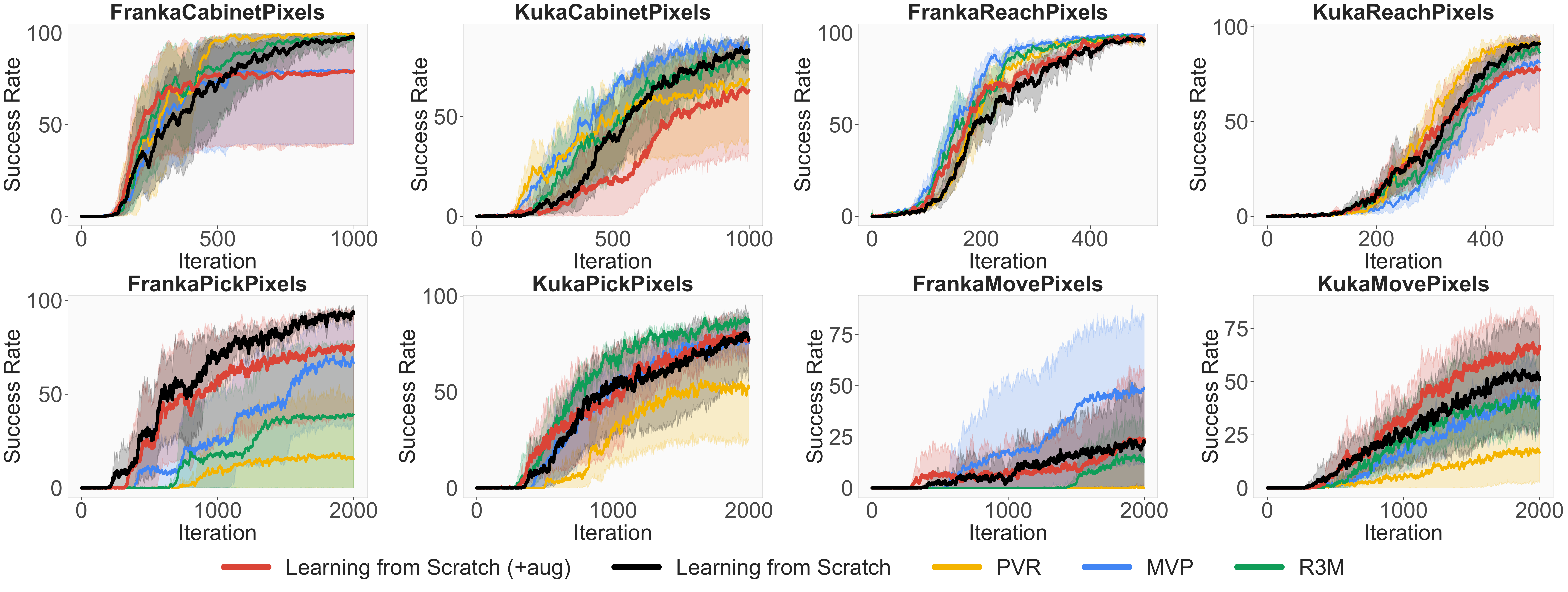}
    \vspace{-0.175in}
    \caption{\textbf{PixMC benchmark.} Success rate of PPO \citep{Schulman2017ProximalPO} agents on the 8 robotic manipulation tasks from PixMC \citep{Xiao2022MaskedVP}. Our proposed LfS baseline performs comparably to the frozen pre-trained visual representations on most tasks. Notably, we also observe that no single pre-trained representation is consistently better across all tasks. Results are averaged across 5 seeds.}
    \label{fig:indv-tasks}
    \vspace{-0.05in}
\end{figure*}
\begin{figure*}[t]
    \centering
    \begin{subfigure}[b]{0.49\textwidth}
        \centering
        \includegraphics[width=\textwidth]{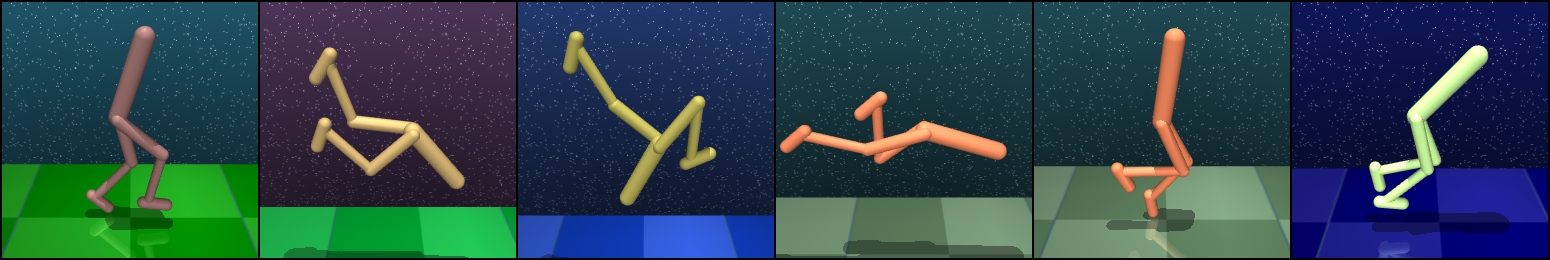}
        \caption{Randomized colors.}
    \end{subfigure}
    \begin{subfigure}[b]{0.49\textwidth}
        \centering
        \includegraphics[width=\textwidth]{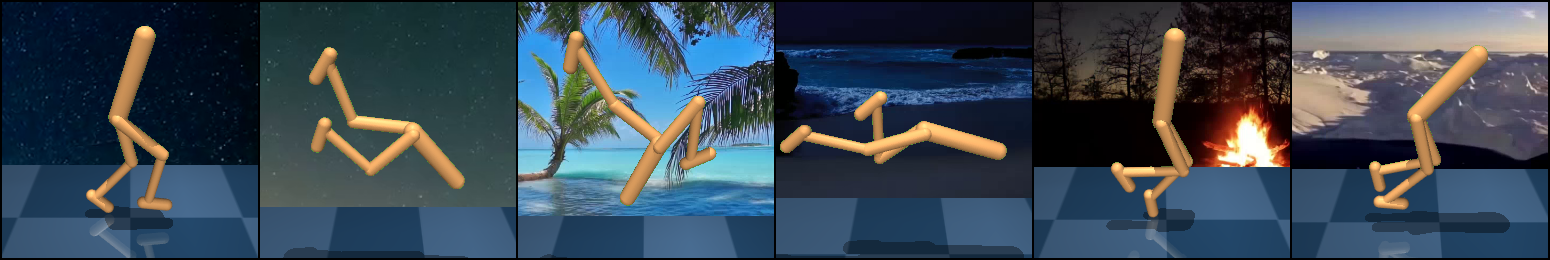}
        \caption{Video backgrounds.}
    \end{subfigure}
    \vspace{-0.15in}
    \caption{\textbf{Evaluation of robustness}. We quantify robustness to visual changes on two test domains from DMControl Generalization Benchmark \citep{hansen2021softda}: \emph{randomized colors} of agent, floor, and background, and dynamic \emph{video backgrounds} sourced from out-of-domain data, corresponding to the \texttt{color\_hard} and \texttt{video\_easy} test domains from the proposed benchmark. Sample environments are visualized. Note that a domain gap remains between augmented observations and test environments.}
    \label{fig:appendix-dmcgb}
\end{figure*}

\vspace{-0.025in}
\section{Results}
\label{sec:experiments-results}
\vspace{-0.025in}
In this section, we present a clear summary of our key experimental results, and defer deeper discussion on the implications of these findings (along with practical guidance for practitioners) to Section \ref{sec:discussion}. Our results are as follows:

$\boldsymbol{-}$ \textbf{Performance comparison.} Our proposed Learning-from-Scratch (LfS) baselines are competitive with (and in some cases \emph{outperform}) recent frozen pre-trained representations for visuo-motor control across a variety of algorithms and domains in both simulation and the real world; see Figure \ref{fig:main-result} and Table \ref{tab:il}. This indicates that, while pre-trained representations have the potential to replace the LfS paradigm in the future, under the set of most widely used metrics, they have yet to exceed the representational power of a \emph{well-designed} LfS method on standard benchmarks for visuo-motor control. This conclusion appears to generalize to real robot tasks with simple visuals.\vspace{0.05in}\\
$\boldsymbol{-}$ \textbf{No free lunch -- yet.} Our results indicate that the efficacy of a frozen pre-trained representation is both \emph{task-dependent} (see Figure \ref{fig:indv-tasks}) and \emph{algorithm-dependent} (see Figure \ref{fig:main-result}): on average, \textcolor{gblue}{$\bullet$}MVP outperforms other pre-trained representations on PixMC for which it was originally proposed, but performs comparably worse on the two other domains, Adroit and DMControl. However, even within a visually consistent benchmark (PixMC), no single representation convincingly comes out on top across tasks, as evidenced by Figure \ref{fig:indv-tasks}. In contrast, our proposed \textcolor{gred}{$\bullet$}\emph{LfS (+aug)} method produces consistently strong results across all settings, presumably due to learning from task-specific data; this hypothesis is supported by our finetuning results, which we return to later.\vspace{0.05in}\\
\begin{figure*}[t]
    \centering
    \includegraphics[width=0.9\linewidth]{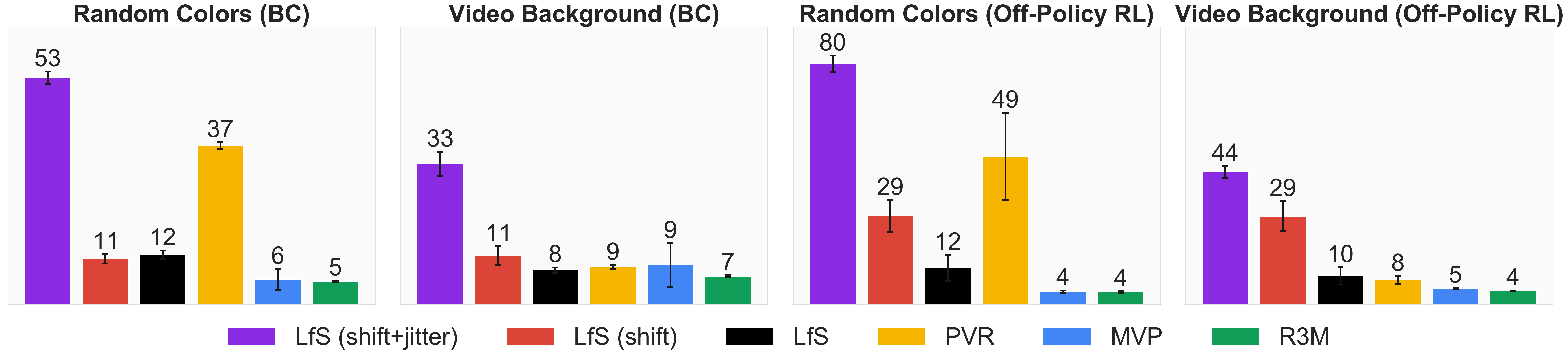}
    \vspace{-0.175in}
    \caption{\textbf{Robustness to visual changes.} Normalized return of methods when transferred to environments with visual changes from the DMControl Generalization Benchmark \citep{hansen2021softda}. We consider two visual changes: randomized colors of agent and scene, as well as dynamic video backgrounds. Following our previous setup, BC results are averages of top-3 evaluations over 100 epochs, and final evaluations are reported for RL results. Mean and $95\%$ confidence intervals over 5 seeds and 4 tasks; we omit \emph{Reacher Easy} since it does not support video backgrounds. LfS with strong augmentation is surprisingly robust compared to frozen pre-trained representations.}
    \label{fig:robustness}
    \vspace{-0.1in}
\end{figure*}
\begin{figure}[t]
    \centering
    \includegraphics[width=0.485\textwidth]{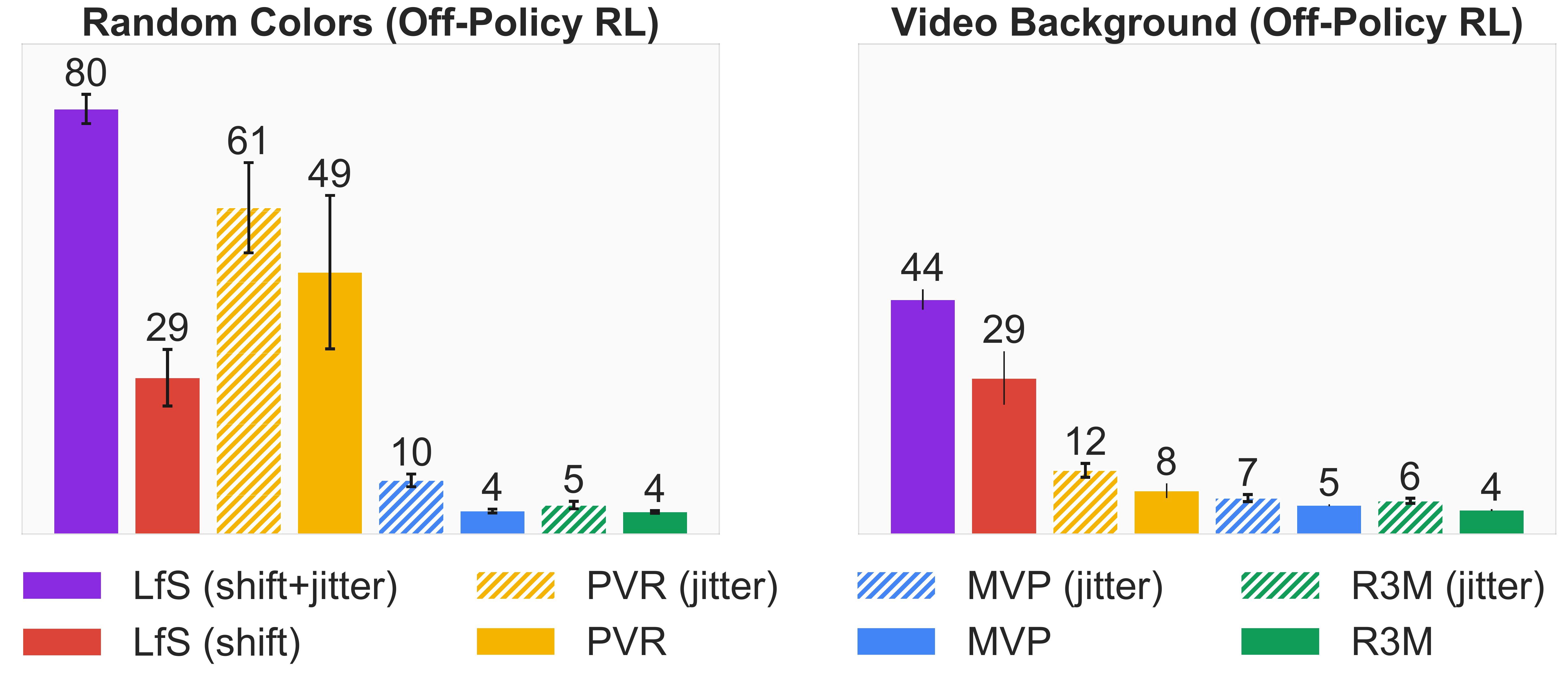}
    \vspace{-0.325in}
    \caption{\textbf{Improving robustness of frozen pre-trained representations with strong augmentation.} Normalized return of methods when transferred to environments with visual changes from the DMControl Generalization Benchmark \citep{hansen2021softda}. We report results both with and without additional \textcolor{gpurple}{$\bullet$}\emph{color jitter} (strong) augmentation during policy learning, and find that applying strong augmentation with a \emph{frozen} representation is ineffective. Mean and $95\%$ confidence intervals over 5 seeds and 4 tasks; we omit \emph{Reacher Easy} since it does not support video backgrounds.}
    \label{fig:robustness-jitter}
    \vspace{-0.15in}
\end{figure}
$\boldsymbol{-}$ \textbf{Visual robustness.} To probe representations for visual robustness, we evaluate trained agents on the DMControl Generalization Benchmark \citep{hansen2021softda}. In this evaluation, agents are trained on the original training environments with no visual variation, and transferred zero-shot to test environments with visual changes. We consider two types of visual changes: \emph{(i)} \emph{random colors} where the colors of agent, background, and floor are randomized, and \emph{(ii)} \emph{video backgrounds} where the background is replaced with a dynamically changing texture from out-of-domain videos; see Figure \ref{fig:appendix-dmcgb} (Appendix) for a visualization of these test environments. Our robustness results are shown in Figure \ref{fig:robustness}. We find that \emph{use of data augmentation is critical to the robustness of learned visual representations} -- both when LfS and when using frozen pre-trained representations. Notably -- in their original formulations -- PVR uses color jitter during pre-training whereas MVP and R3M do not. We compare robustness of these pre-trained representations with LfS using both \textcolor{gred}{$\bullet$}\emph{random shift} augmentation and additionally \textcolor{gpurple}{$\bullet$}\emph{color jitter}. We find that \emph{(1)} pre-trained representations that do \emph{not} use color jitter augmentation during pre-training (\textcolor{gblue}{$\bullet$}MVP, \textcolor{ggreen}{$\bullet$}R3M) are not more robust than their LfS counterpart to visual changes, but \emph{(2)} strong augmentations such as \textcolor{gpurple}{$\bullet$}\emph{color jitter} improves robustness of \emph{both} LfS and pre-trained representations (\textcolor{gyellow}{$\bullet$}PVR; applied during pre-training) significantly. For completeness, we also evaluate LfS with a different choice of strong augmentation: \textcolor{gray}{$\bullet$}\emph{random overlay} that interpolates between observations and randomly sampled images from an out-of-domain dataset, popularized by \citet{hansen2021softda}; results are shown in Table \ref{tab:overlay}. Consistent with prior work, we find that choice of augmentation influences robustness to different visual changes, but that \emph{either choice beats the best pre-trained representation} that we consider.\vspace{0.05in}\\
\begin{table}[t]
    \centering
    \vspace*{-9pt}
    \captionof{table}{\textbf{Choice of augmentation matters.} Mean normalized return of BC policies when transferred to environments with visual changes from the DMControl Generalization Benchmark \citep{hansen2021softda}. We here consider LfS with two distinct choices of strong data augmentation: \emph{color jitter} as in Figure \ref{fig:robustness}, and \emph{random overlay} originally proposed by \citet{hansen2021softda}; these augmentations are in addition to random image shifts. For completeness, we also include our \textbf{best} result obtained with a pre-trained representation, \emph{i.e.}, we report $\max\{\textrm{PVR}, \textrm{MVP}, \textrm{R3M}\}$.}
    \vspace{0.075in}
    \label{tab:overlay}
    \resizebox{0.44\textwidth}{!}{%
    \begin{tabular}[b]{lcccc}
    \toprule
    \textbf{~~~~~~~~~~~~~~~~Method} & \textcolor{gpurple}{$\bullet$} LfS & \textcolor{gray}{$\bullet$} LfS & \textbf{Best}
    \\
    \textbf{Test set} & (\emph{jitter}) & (\emph{overlay}) & pre-training
    \\ \midrule
    Random colors         & $\mathbf{53.1\scriptstyle{\pm 1.6}}$   & $39.3\scriptstyle{\pm 0.2}$ & $37.2\scriptstyle{\pm 0.9}$\\
    Video background    & $33.0\scriptstyle{\pm 3.2}$   & $\mathbf{46.6\scriptstyle{\pm 1.3}}$   & $9.2\scriptstyle{\pm 5.8}$ \\ \bottomrule
    \end{tabular}
    }
    \vspace{-0.15in}
\end{table}
$\boldsymbol{-}$ \textbf{Finetuning a pretrained representation.} To help narrow the scope of our comparison, our study primarily considers frozen visual representations following their original proposals, \emph{i.e.}, neither PVR, MVP, or R3M finetune their representations on in-domain data. However, there is \emph{some} existing evidence that in-domain finetuning of pre-trained representations can be beneficial \citep{Wang2022VRL3AD, Ze2022rl3d, xu2022xtra}. For completeness, we also conduct a set of finetuning experiments, where pre-trained representations (\textcolor{gyellow}{$\bullet$}PVR, \textcolor{gblue}{$\bullet$}MVP, \textcolor{ggreen}{$\bullet$}R3M) are finetuned on demonstration data from Adroit using the task-centric behavior cloning objective. Results for this experiment are shown in Table \ref{tab:finetuning}. Interestingly, we find that finetuned representations can improve over both their frozen counterparts and our \textcolor{gred}{$\bullet$}\emph{LfS (+aug)} approach, but \emph{only when also using data augmentation} (random shift) during finetuning. This observation indicates that data augmentation is critical to performance when learning on a small (by comparison) in-domain dataset, \emph{regardless} of whether the representation is learned from scratch or finetuned. We are -- to the best of our knowledge -- the first to make this observation, and conjecture that this discrepancy in performance is due to a domain gap between out-of-domain training data and in-domain data. Given that our finetuning experiments are in simulation whereas the pre-training data consists of real-world images, we dub this phenomenon the \emph{real-to-sim} gap. However, our real robot results also indicate that this gap persists to some extent even when evaluating in the real world. Lastly, we also find that ResNet-based representations (\textcolor{gyellow}{$\bullet$}PVR, \textcolor{ggreen}{$\bullet$}R3M) are easier to finetune than ViT (\textcolor{gblue}{$\bullet$}MVP), presumably due to known optimization challenges in ViTs \citep{Dosovitskiy2021AnII, Chen2021WhenVT, Hansen2021StabilizingDQ}. We consider frozen visual representations in the remainder of our experiments, but provide further discussion on the potential implications of this observation in Section \ref{sec:discussion}. \vspace{0.05in}\\
\begin{table*}[t]
    \centering
    \vspace*{-9pt}
    \captionof{table}{\textbf{Wall-time} of methods learning from scratch vs. using a pre-trained visual representation. For the latter, we report $\min\{\textrm{PVR}, \textrm{MVP}, \textrm{R3M}\}$ for a fair comparison. While LfS generally leads to better downstream task performance, using a \textbf{frozen} pre-trained representation can reduce computational cost substantially, especially during the training process. $\downarrow$ Lower is better.}
    \vspace{0.05in}
    \label{tab:wall-time}
    \resizebox{0.75\textwidth}{!}{%
    \begin{tabular}[b]{lcc|cc|cc}
    & \multicolumn{4}{c}{\textbf{Behavior Cloning}} & \multicolumn{2}{c}{\textbf{Reinforcement Learning}} \\
    & \multicolumn{2}{c}{Training (s/iteration)} & \multicolumn{2}{c}{Inference (s/episode)} & s/$1$k frames & s/iteration \\
    \toprule
    \textbf{Method\textbackslash Setting} & Adroit & DMControl & Adroit & DMControl & DrQ-v2 & PPO \\ \midrule
    \textcolor{gred}{$\bullet$} LfS (+aug)  &   $0.263$ & $0.270$  &  $\mathbf{1.61}$ & $\mathbf{3.81}$ & $\mathbf{10.20}$  & $19.40$ \\
    \textbf{Fastest} pre-training & $\mathbf{0.003}$ & $\mathbf{0.006}$ & $2.66$ & $11.00$ & $13.00$ & $\mathbf{11.90}$\\
    \bottomrule
    \end{tabular}
    }
    \vspace{-0.125in}
\end{table*}
\begin{table}[t]
    \centering
    \vspace*{-6pt}
    \captionof{table}{\textbf{Finetuning pre-trained representations with BC.} Success rate for each method across 5 seeds and all Adroit tasks. \emph{Finetuned} denotes whether a representation has been finetuned on task data, and \emph{data aug} denotes whether random image shift augmentation is applied during finetuning. We find that finetuning ResNet-based representations (PVR, R3M) on task data improves over frozen representations and even outperforms LfS (+aug), but \emph{only when using data augmentation during finetuning}. We are -- to the best of our knowledge -- the first to make this observation.}
    \vspace{0.075in}
    \label{tab:finetuning}
    \resizebox{0.475\textwidth}{!}{%
    \begin{tabular}[b]{lccccc}
    \toprule
    \textbf{Method} & Finetuned & Data aug & Success (\%) & Change
    \\ \midrule
    \textcolor{gyellow}{$\bullet$} \underline{PVR} & \textcolor{gred}{\XSolidBrush} & \textcolor{gred}{\XSolidBrush} & $52.9\scriptstyle{\pm2.1}$ & $-$ \\
    & \textcolor{ggreen}{\Checkmark} & \textcolor{gred}{\XSolidBrush} & $50.5\scriptstyle{\pm7.7}$ & \textcolor{gred}{$\mathbf{-2.4}$} \\
    & \textcolor{ggreen}{\Checkmark} & \textcolor{ggreen}{\Checkmark} & $65.0\scriptstyle{\pm0.0}$ & \textcolor{ggreen}{$\mathbf{+12.1}$} \\ \midrule
    \textcolor{gblue}{$\bullet$} \underline{MVP} & \textcolor{gred}{\XSolidBrush} & \textcolor{gred}{\XSolidBrush} & $44.0\scriptstyle{\pm2.2}$ & $-$ \\
    & \textcolor{ggreen}{\Checkmark} & \textcolor{gred}{\XSolidBrush} & $18.7\scriptstyle{\pm1.9}$ & \textcolor{gred}{$\mathbf{-25.3}$} \\
    & \textcolor{ggreen}{\Checkmark} & \textcolor{ggreen}{\Checkmark} & $31.1\scriptstyle{\pm2.7}$ & \textcolor{gred}{$\mathbf{-12.9}$} \\ \midrule
    \textcolor{ggreen}{$\bullet$} \underline{R3M} & \textcolor{gred}{\XSolidBrush} & \textcolor{gred}{\XSolidBrush} & $64.0\scriptstyle{\pm2.8}$ & $-$ \\
    & \textcolor{ggreen}{\Checkmark} & \textcolor{gred}{\XSolidBrush} & $55.5\scriptstyle{\pm8.9}$ & \textcolor{gred}{$\mathbf{-8.5}$} \\
    & \textcolor{ggreen}{\Checkmark} & \textcolor{ggreen}{\Checkmark} & $\mathbf{80.5\scriptstyle{\pm2.1}}$ & \textcolor{ggreen}{$\mathbf{+16.5}$} \\ \midrule
    \textcolor{black}{$\bullet$} LfS & \textcolor{ggreen}{\Checkmark} & \textcolor{gred}{\XSolidBrush} & $19.4\scriptstyle{\pm3.8}$ & $-$ \\
    \textcolor{gred}{$\bullet$} LfS \emph{(+aug)} & \textcolor{ggreen}{\Checkmark} & \textcolor{ggreen}{\Checkmark} & $63.1\scriptstyle{\pm6.4}$ & \textcolor{ggreen}{$\mathbf{+43.7}$} \\ \bottomrule
    \end{tabular}
    }
    \vspace{-0.15in}
\end{table}
\begin{figure}[t]
    \centering
    \includegraphics[width=0.475\textwidth]{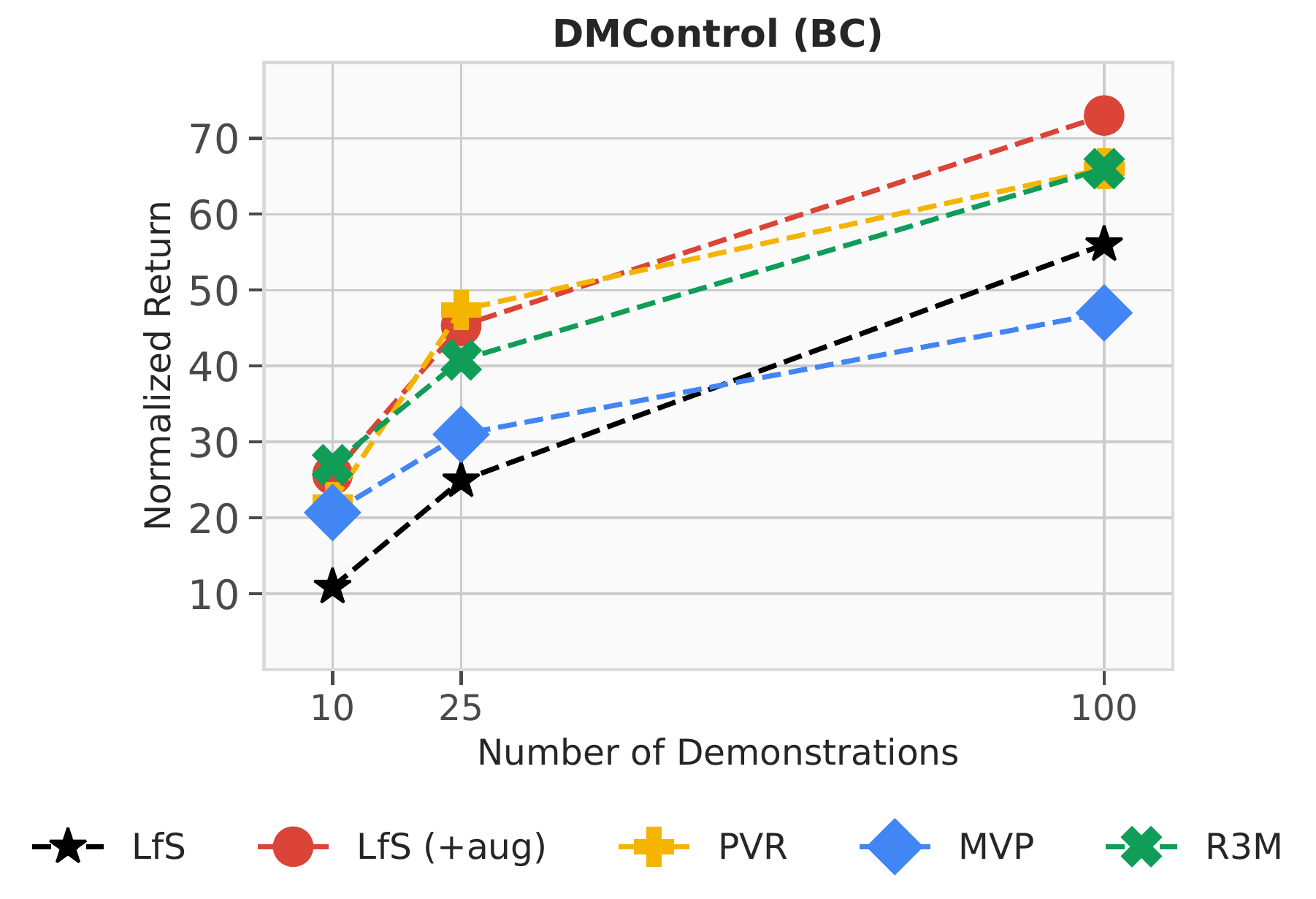}
    \vspace{-0.15in}
    \caption{\textbf{Data efficiency.} Normalized return of behavior cloning policies as a function of the number of demonstrations. Results are averaged across all of our DMControl tasks and 3 seeds. We find that a larger amount of demonstrations (100) favors LfS, whereas pre-trained representations fare better in the very low-data regime.}
    \label{fig:scaling}
    \vspace{-0.15in}
\end{figure}
$\boldsymbol{-}$ \textbf{Data efficiency.} A common argument in favor of (frozen) pre-trained representations is that they might require less task-specific data to learn a good policy \citep{Parisi2022TheUE, Xiao2022MaskedVP, Nair2022R3MAU}. To test this hypothesis, we train BC policies with a variable number of demonstrations (10, 25, 100) for both our improved LfS baseline and the three frozen pre-trained representations; following the experimental setup of PVR, we use 100 demonstrations in the remainder of our BC experiments. We report the results of this experiment in Figure \ref{fig:scaling}. Our results indicate that a larger number of demonstrations (100) generally favors LfS methods, whereas frozen pre-trained representations fare marginally better in the very low-data regime (10). However, policy performance degrades quickly with a decrease in available demonstrations, which suggests that the primary performance bottleneck is in policy learning rather than visual representation learning. As discussed in Section \ref{sec:discussion}, this may in part be due to current benchmarks being visually simple. We predict that this observation might not continue to hold true as new simulation benchmarks with more complex visuals are developed.\vspace{0.05in}\\
$\boldsymbol{-}$ \textbf{Computational cost.} Our results so far have focused on downstream task performance, \emph{i.e.}, success rate or return in various settings. However, frozen pre-trained representations already demonstrate significant gains along an often-neglected axis: \emph{wall-time}. Training and inference speeds are shown in Table \ref{tab:wall-time}. We find that BC policy updates are at least an order of magnitude faster using frozen pre-trained representations compared to LfS, as we can embed and cache features for the entire dataset in a few forward passes. However, inference speed generally favors LfS due to their smaller visual backbones, which is particularly important for real robot applications. Since RL training interleaves learning and inference (data collection), wall-times are more balanced in this setting. {We do not factor in the cost of learning a pre-trained representation, since it is a one-time cost, and the representations can be reused across tasks}.

\section{Discussion}
\label{sec:discussion}
\vspace{-0.05in}
We have shown that a carefully designed LfS baseline is competitive with frozen pre-trained representations across a variety of algorithm classes, domains, and metrics. While this is the current conclusion, we remain optimistic that results will be skewed in favor of pre-trained representations as the paradigm matures. At present, we find that the main benefit of a frozen pre-trained representation is the reduced training cost that comes with its \emph{universality} -- a single representation can be reused across tasks. Achieving a performance edge while maintaining universality will thus be critical to the adoption of this new paradigm. Our experiments indicate that pre-trained representations benefit from finetuning on task-specific data (when coupled with use of data augmentation), combining elements of pre-training and LfS. However, finetuning large visual backbones presents optimization challenges (\emph{e.g.}, catastrophical forgetting and instability), and can be costly. In the following, we share our vision for the future of pre-training research for control, which we hope can inspire further research in the area. However, we remark that this vision -- while being informed by findings in this work -- is ultimately a conjecture.

$\boldsymbol{-}$ \textbf{A benchmark perspective.} Our first conjecture is that current benchmark tasks are not well suited to reap the benefits of pre-trained representations. Historically, the majority of visual RL benchmarks have been repurposed from existing environments that were originally proposed for RL from ground-truth (state) information, with little emphasis on visual complexity, variation, and realism. Furthermore, there have historically been strong emphasis on single-task learning, where limited semantic information is required. In such (visually) simple settings, it is perhaps not surprising that learning a representation from scratch on in-domain data is sufficient and oftentimes better than a general-purpose representation trained solely on out-of-domain ImageNet or human interaction data. We predict that pre-trained representations (frozen and finetuned alike) will fare better as new benchmarks with these properties -- visual complexity, variation, realism, and multi-task learning -- are developed. To this end, we view evaluation of policies in the real world -- such as those shown in Figure \ref{fig:real-robot-tasks} -- as a step in the right direction. However, most contemporary real robot setups (for which ours is no exception) still leave much to be desired in terms of visual complexity and variation compared to the diversity of the pre-training data leveraged in research on pre-trained representations for control.

$\boldsymbol{-}$ \textbf{A domain gap perspective.} Our second conjecture is directly informed by our experiments. We observe that, while LfS consistently outperforms current frozen pre-trained representations, finetuning on in-domain data (with the addition of data augmentation) results in representations that achieve better downstream task performance and are more robust to visual variations compared to their frozen counterparts and in some cases even LfS. This important result suggests that this discrepancy is due to a large domain gap between pre-training data and in-domain data. Given that current pre-trained representations are learned from out-of-domain ImageNet or human interaction data (\emph{i.e.}, real-world data) and predominantly tested in simulated robot environments (distinctly different domains), it is perhaps not surprising that finetuning representations on a small amount of in-domain data can lead to markedly better downstream task performance. We dub this domain gap the \emph{real-to-sim} gap, although we empirically find that the problem persists to some extent even in real robot experiments. We recommend future work to either \emph{(1)} pre-train on data that better reflect the data distribution of downstream tasks (\emph{e.g.}, by training on simulation data or real-world robot data, or by evaluating policies in real world scenes that more closely match those present in existing pre-training datasets), or \emph{(2)} finetune on a small in-domain dataset using, \emph{e.g.}, a task-centric objective such as BC (if demonstrations are available) or RL (online interaction). While addressing the domain gap by collecting a new dataset for pre-training can be costly, it is relatively easy to finetune current models on a small in-domain dataset; this is reminiscent to the current trend of aligning large language models by finetuning on small curated datasets \cite{alpaca}. Lastly, our experiments demonstrate that data augmentation is absolutely critical to learning strong, robust representations in all stages of training. To the best of our knowledge, we are the first work on pre-trained representations for control to make this discovery. In comparison to the refined training recipes in computer vision literature, training recipes for visuo-motor control are still relatively underexplored. We predict that -- as pre-processing and data augmentation training recipes for visuo-motor control mature -- we will see a series of increasingly robust pre-trained representations emerge. We encourage further research in all of these directions, and hope that our LfS baselines will help accurately benchmark progress in this area.

\section{Related Work}
\label{sec:related-work}
\textbf{Pre-training.} Representation learning via supervised/self-supervised/unsupervised pre-training on large-scale datasets has emerged as a powerful paradigm in areas such as computer vision \citep{Doersch2015UnsupervisedVR, He2020MomentumCF, Oord2018RepresentationLW, alayrac2022flamingo} and natural language processing \citep{Devlin2019BERTPO, Brown2020LanguageMA, radford2021learning, chowdhery2022palm}, where large datasets are available. While pre-trained representations can be finetuned to solve various downstream tasks, it may be prohibitively expensive to do so, and representations are therefore commonly used as-is, \emph{i.e.}, with \emph{frozen} weights. We reflect on recent progress and challenges when leveraging pre-trained visual representations for control, which is an emerging and comparably underexplored application area of such representations.

\textbf{Pre-trained representations for control.} Multiple works have explored learning control policies with visual representations pre-trained on large external datasets \citep{Shah2021RRLRA, Parisi2022TheUE, Nair2022R3MAU, Xiao2022MaskedVP, Wang2022VRL3AD, Ze2022rl3d, yuan2022pretrained, xu2022xtra, rt12022arxiv}. In particular, PVR \citep{Parisi2022TheUE} and R3M \citep{Nair2022R3MAU} propose to learn policies by behavior cloning using pre-trained representations; PVR fuses features from several layers of a ResNet50 learned by MoCo-v2 \citep{He2020MomentumCF}, and R3M \citep{Nair2022R3MAU} learns a representation using a time-contrastive objective on ego-centric human videos. MVP \citep{Xiao2022MaskedVP} learns a policy with PPO \citep{Schulman2017ProximalPO} and uses a pre-trained visual encoder for feature extraction in addition to proprioceptive state information; the pre-trained representation is an MAE \citep{He2022MaskedAA} trained on frames from diverse human videos. We show that our improved LfS baseline remains competitive with (frozen) pre-trained representations, but also find that an equally carefully designed \emph{finetuning} procedure of pre-trained representations can outperform LfS in some cases.

\textbf{Data augmentation in RL.} Numerous recent studies demonstrate the effectiveness of data augmentation in visual RL \citep{Lee2019ASR, Laskin2020ReinforcementLW, raileanu2020automatic, Kostrikov2021ImageAI, yarats2021mastering, hansen2021softda, Hansen2021StabilizingDQ, ma2022comprehensive, Hansen2022tdmpc}. For example, \citet{Lee2019ASR, hansen2021softda} show that strong data augmentation can greatly improve the visual robustness and generalization of RL policies. Domain randomization \citep{Tobin_2017, pinto2017asymmetric}, a closely related idea, has similarly been shown to improve generalization and sim-to-real transfer. \citet{Laskin2020ReinforcementLW} conducts a comprehensive study on data augmentations for RL, and finds that random crops can lead to significant gains in both sample-efficiency and asymptotic performance. Finally, \citet{Kostrikov2021ImageAI, yarats2021mastering} propose a simple \emph{random shift} augmentation that further improves over random crop augmentation in the context of visual off-policy RL; we apply this augmentation in all of our \textcolor{gred}{$\bullet$}\emph{LfS (+aug)} experiments. Our study confirms the observations of prior work, and shows that the resulting LfS baselines remain competitive with frozen pre-trained representations trained on large-scale out-of-domain datasets.

\section{Conclusion}
\label{sec:conclusion}
To conclude, we reiterate the main takeaways of our study:\vspace{0.075in}\\
$\boldsymbol{-}$ A \emph{carefully designed} LfS approach remains competitive with frozen pre-trained representations across a variety of algorithms, task domains, and evaluation metrics.\vspace{0.05in}\\
$\boldsymbol{-}$ At this time, no single frozen pre-trained representation is consistently better across all tasks.\vspace{0.05in}\\
$\boldsymbol{-}$ Finetuning pre-trained representations on task-specific data leads to significant improvements in performance (when also using data augmentation during finetuning), even surpassing the performance of \textcolor{gred}{$\bullet$}LfS \emph{(+aug)} in some cases.\vspace{0.05in}\\
$\boldsymbol{-}$ LfS with strong data augmentation (\textcolor{gpurple}{$\bullet$}\emph{color jitter}) outperforms frozen pre-trained representations by a large margin on visual robustness benchmarks. However, adding strong data augmentation to pre-training and policy learning pipelines consistently improves their robustness.\vspace{0.05in}\\
$\boldsymbol{-}$ Pre-trained representations fare slightly better than LfS approaches in the very low-data regime, but our experiments indicate that policy learning might be a bigger bottleneck when data is limited.\vspace{0.05in}\\
$\boldsymbol{-}$ Using \emph{frozen} pre-trained representations can lead to significant improvements in training wall-time, at the expense of slower inference compared to our smaller LfS backbones. Since RL training interleaves learning and inference (data collection), wall-times are more balanced in this setting.\vspace{0.05in}\\
$\boldsymbol{-}$ We remain optimistic about the future of pre-trained representations for visuo-motor control, and hope that our strong LfS baselines will help accurately benchmark progress in the area.

\paragraph{Acknowledgements} Experiments were divided equally between NH, ZY, YZ, and TM. AR, HS, HX, and XW advised equally. All authors contributed to writing. The authors would like to thank Ziyan Xiong for their help with conducting real robot experiments for the camera-ready version of the paper. Prof. XW's lab is supported, in part, by an Amazon Research Award and gifts from Qualcomm.

\bibliography{main}
\bibliographystyle{icml2023}

\newpage
\appendix
\onecolumn

\section{Task Descriptions}
\label{sec:appendix-tasks}
We conduct experiments on three different task domains in simulation -- Adroit \citep{Rajeswaran-RSS-18}, DMControl \citep{deepmindcontrolsuite2018}, and PixMC \citep{Xiao2022MaskedVP}-- used in prior work on pre-training for visuo-motor control, as well as a real robot setup. PVR \citep{Parisi2022TheUE} experiments with Adroit and DMControl, MVP \citep{Xiao2022MaskedVP} proposed the PixMC benchmark, and R3M \citep{Nair2022R3MAU} experiments with Adroit (among others). To make our study more self-contained, we include a detailed description of each task below.

\subsection{Adroit}
\label{sec:appendix-adroit}
Following PVR, we consider two tasks from the Adroit domain: \textit{pen} and \textit{relocate}, which represent the two most challenging tasks from this task domain. The two Adroit tasks are  goal-conditioned dexterous manipulation tasks with goals rendered visually in a 3D scene, as shown in Figure \ref{fig:tasks} \emph{(left)}. The robot hand has 24 degrees of freedom (DoF). We describe each task as follows:

\begin{itemize}
    \item \textit{Pen} ($\mathcal{A} \in \mathbb{R}^{18}$). A blue pen is initialized in the palm of the dexterous robot hand. The task is to reorient the pen in-hand to bring it to a desired orientation, which is visualized as a transparent pen floating next to the hand. The agent controls all joints but its wrist is locked and cannot move in 3D space.
    \item \textit{Relocate} ($\mathcal{A} \in \mathbb{R}^{21}$). A blue ball is initialized at a random location on a table. The task is to pick up the ball using the dexterous robot hand, and move it to a desired (randomly selected) location in 3D space, which is visualized as a transparent green ball. The agent controls all joints, as well as the wrist which can move freely in 3D space.
\end{itemize}

We refer to \citet{Rajeswaran-RSS-18} for additional task details.

\subsection{DMControl}
\label{sec:appendix-dmcontrol}
Following PVR, we consider five tasks from the DMControl suite: \textit{Finger Spin}, \textit{Reacher Easy}, \textit{Cheetah Run}, \textit{Walker Stand}, and \textit{Walker Walk}, which represent continuous control tasks of varying difficulty. These tasks vary in embodiment, objective, action space, and reward type. Two of the DMControl tasks (\emph{Finger Spin} and \emph{Walker Walk}) are visualized in Figure \ref{fig:tasks} \emph{(center left)}. We describe each task as follows:

\begin{itemize}
    \item \textit{Finger Spin} ($\mathcal{A} \in \mathbb{R}^{2}$). A simple manipulation task with a planar 3 DoF finger. The task is to continuously spin a free-floating body at high velocity. There is a positive reward of $+1$ for each timestep that the body is spinning and $0$ otherwise.
    \item \textit{Reacher Easy} ($\mathcal{A} \in \mathbb{R}^{2}$). A simple manipulation task with a planar 3 DoF finger. The task is to move the fingertip to a randomly selected location in 2D space. There is a positive reward of $+1$ for each timestep that the fingertip is near the target and $0$ otherwise.
    \item \textit{Cheetah Run} ($\mathcal{A} \in \mathbb{R}^{6}$). A locomotion task with a planar cheetah embodiment. The task is to run forward at high velocity until the end of the episode. There is a dense (shaped) reward that varies with forward velocity and positioning of joints.
    \item \textit{Walker Stand} ($\mathcal{A} \in \mathbb{R}^{6}$). A locomotion task with a planar Walker embodiment. The task is to stand up until the end of the episode. There is a dense (shaped) reward that varies with the positioning of joints.
    \item \textit{Walker Walk} ($\mathcal{A} \in \mathbb{R}^{6}$). A locomotion task with a planar Walker embodiment. The task is to walk forward at medium velocity until the end of the episode. There is a dense (shaped) reward that varies with forward velocity and positioning of joints.
\end{itemize}

We refer to \citet{deepmindcontrolsuite2018} for additional task details.

\subsection{PixMC}
\label{sec:appendix-pixmc}
Following MVP, we consider all eight tasks from the proposed PixMC benchmark. These tasks consist of four robot manipulation tasks (\emph{Cabinet}, \emph{Pick}, \emph{Move}, and \emph{Reach}) across two 7-DoF robots (\emph{Franka} Emika and \emph{Kuka} LBR iiwa) that use a mounted parallel jaw gripper and multi-finger hand, respectively. Observations are captured by a wrist-mounted camera. Besides embodiment and action space, these tasks also vary in interaction type and difficulty, and there is variability in objects and locations between each episode. Two of the PixMC tasks (\emph{Kuka Pick} and \emph{Franka Move}) are shown in Figure \ref{fig:tasks} \emph{(center right)}. We describe each task as follows:

\begin{itemize}
    \item \textit{Reach} (Franka: $\mathcal{A} \in \mathbb{R}^{9}$, Kuka: $\mathcal{A} \in \mathbb{R}^{23}$). A simple manipulation task. The task is to reach a randomly selected location in 3D space with the end-effector. There is a dense (shaped) reward.
    \item \textit{Cabinet} (Franka: $\mathcal{A} \in \mathbb{R}^{9}$, Kuka: $\mathcal{A} \in \mathbb{R}^{23}$). A complex articulated object manipulation task. The task is open the top drawer of a free-standing cabinet. There is a dense (shaped) reward.
    \item \textit{Pick} (Franka: $\mathcal{A} \in \mathbb{R}^{9}$, Kuka: $\mathcal{A} \in \mathbb{R}^{23}$). An object manipulation task. The task is to pick up a randomly initialized object from the table, and hold it above a certain height threshold. There is a dense (shaped) reward.
    \item \textit{Move} (Franka: $\mathcal{A} \in \mathbb{R}^{9}$, Kuka: $\mathcal{A} \in \mathbb{R}^{23}$). An object manipulation task. The task is to move a randomly initialized object to a different location. There is a dense (shaped) reward.
\end{itemize}

We refer to \citet{Xiao2022MaskedVP} for additional task details.

\subsection{Real robot}
\label{sec:appendix-real-robot}
In addition to our three simulation domains, we also consider two manipulation tasks on a real robot: \textit{reach} and \textit{pick}, which resemble the two PixMC tasks of the same name. Our experimental setup roughly mimics that of R3M. The agent controls a 7-DoF xArm 7 robot with a jaw gripper using positional control, and visual observations are captured by a static third-person Intel RealSense camera. We randomize object configuration between each episode. To minimize human bias in evaluation, we use a manually designed success detector to determine whether a given trial is successful. The two real robot tasks are visualized in Figure \ref{fig:tasks} \emph{(right)}. We describe each task as follows:

\begin{itemize}
    \item \textit{Reach} ($\mathcal{A} \in \mathbb{R}^{3}$). A blue target is initialized at a random location within the robot workspace. The task is to move the end-effector (grasping a red object) to the target location. This task is therefore goal-conditioned. The agent controls the end-effector using positional control and its gripper is locked. We evaluate success based on the distance between the end-effector and goal at the end of a trial.
    \item \textit{Pick} ($\mathcal{A} \in \mathbb{R}^{4}$). A red octagonal prism is initialized at a random location within the robot workspace. The task is to pick up the object using the gripper, and lift it above a predefined height threshold. The agent controls both the end-effector and gripper using positional control. We evaluate success based on a binary threshold on the end-effector height (assuming that the object is successfully grasped) at the end of a trial.
\end{itemize}

\section{Implementation Details}
\label{sec:appendix-implementation-details}
We provide further implementation details on our improved LfS baselines in the following. For simplicity, we separate the implementation details by algorithm class, but remark that all details not pertaining to the changes that we make to the LfS baselines (shallow ConvNet encoder and data augmentation) are kept identical to prior work to ensure a fair comparison, \emph{i.e.}, we do \emph{not} modify the experimental setup nor hyperparameters. Our code is made available at {\small\url{https://github.com/gemcollector/learning-from-scratch}}.

\subsection{Behavior Cloning}
\label{sec:appendix-bc}
\begin{figure}[t]
    \centering
    \includegraphics[width=0.54\textwidth]{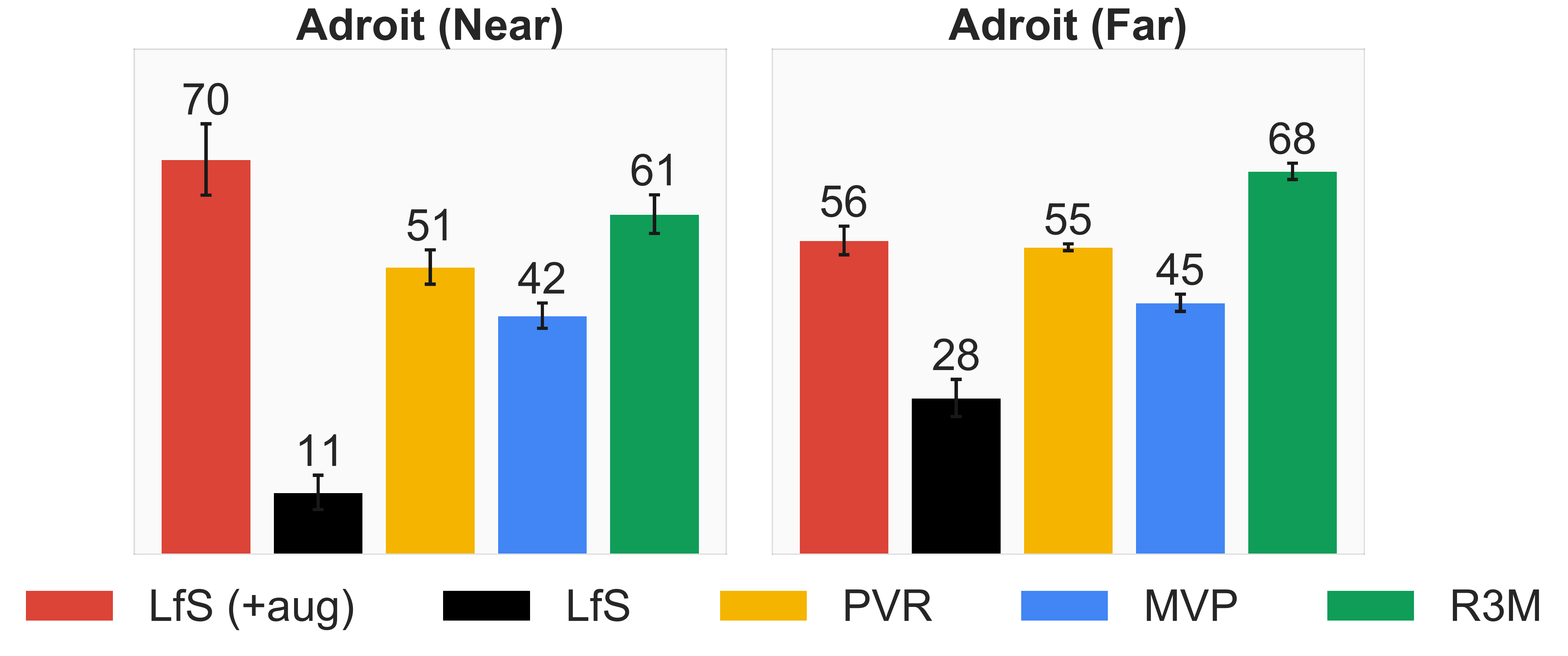}
    \includegraphics[width=0.21\textwidth]{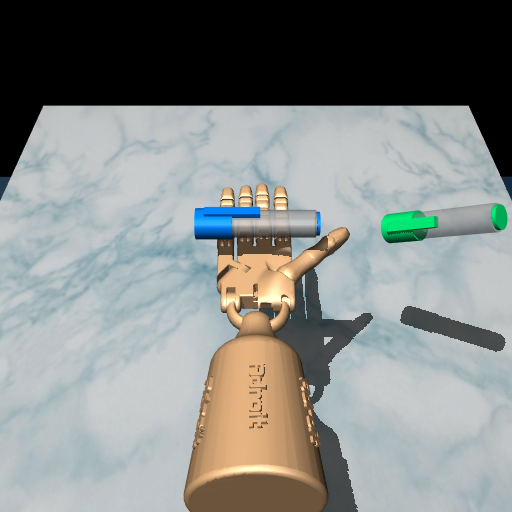}
     \includegraphics[width=0.21\textwidth]
     {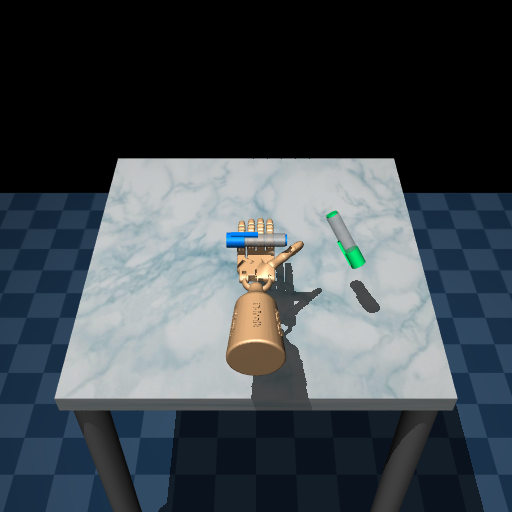}
      
    {\small~~~~~~~~~~~~~~~~(a) Results for individual views.~~~~~~~~~~~~~~~~~~~~~~~~~~~~~~~~~~~~~~~~~~~~~~~(b) Near view.~~~~~~~~~~~~~~~~~~~~~~~(c) Far view.}
    \vspace{-0.075in}
    \caption{\textbf{Results for individual camera views in Adroit.} To improve reliability of our results, we report the average success rate over two camera views for Adroit: \emph{Near} (\texttt{fixed}) and \emph{Far} (\texttt{vil\_camera}). PVR \citep{Parisi2022TheUE} reports results for the \emph{Far} view only. All numbers are means across two tasks and 5 seeds. We find that pre-trained representations benefit more from the farther view, whereas LfS benefits more from the near view.}
    \label{fig:appendix-adroit-cameras}
    \vspace{-0.1in}
\end{figure}

We closely follow the implementation of PVR \citep{Parisi2022TheUE} for our LfS baseline in both the Adroit, DMControl, and real robot task domains. Specifically, the network consists of an encoder:
{\small
\begin{codesnippet}
(0): Conv2d(3, 32, kernel_size=(3, 3), stride=(2, 2), padding=(1, 1))
(1): BatchNorm2d(32, eps=1e-05, momentum=0.1, affine=True, track_running_stats=True)
(2): ReLU()
(3): Conv2d(32, 32, kernel_size=(3, 3), stride=(2, 2), padding=(1, 1))
(4): BatchNorm2d(32, eps=1e-05, momentum=0.1, affine=True, track_running_stats=True)
(5): ReLU()
(6): Conv2d(32, 32, kernel_size=(3, 3), stride=(2, 2), padding=(1, 1))
(7): BatchNorm2d(32, eps=1e-05, momentum=0.1, affine=True, track_running_stats=True)
(8): ReLU()
(9): Conv2d(32, 32, kernel_size=(3, 3), stride=(2, 2), padding=(1, 1))
(10): BatchNorm2d(32, eps=1e-05, momentum=0.1, affine=True, track_running_stats=True)
(11): ReLU()
(12): Conv2d(32, 32, kernel_size=(3, 3), stride=(2, 2), padding=(1, 1))
(13): BatchNorm2d(32, eps=1e-05, momentum=0.1, affine=True, track_running_stats=True)
(14): ReLU()
(15): Flatten(start_dim=1, end_dim=-1)

\end{codesnippet}
}
and a policy head:
{\small
\begin{codesnippet}
(0): Linear(in_features=Z, out_features=256, bias=True)
(1): ReLU()
(2): Linear(in_features=256, out_features=256, bias=True)
(3): ReLU()
(4): Linear(in_features=256, out_features=256, bias=True)
(5): ReLU()
(6): Linear(in_features=256, out_features=A, bias=True)

\end{codesnippet}
}
where \texttt{Z} and \texttt{A} denote the dimensionality of the encoder output and action space, respectively. As in PVR, the encoder encodes images in a stack individually and fuses features using Flare \citep{Shang2021ReinforcementLW} in our simulation experiments. In our real robot experiments, we find it sufficient to use a single frame. The policy has an additional 1D BatchNorm layer at the beginning when using pre-trained representations. We apply random shift augmentation to inputs (a stack of $256\times256$ RGB images with no access to state information) using a padding of $12$ to keep the padding-to-image ratio consistent with its original proposal. All augmentations are applied to the stack consistently across time (if applicable). We use 100 expert demonstrations for each task in simulation, and 10-20 demonstrations in the real world depending on the task (reach: 10, pick: 20). Following PVR, demonstrations are collected using oracle (state-based) DAPG policies in Adroit \citep{Rajeswaran-RSS-18} and oracle DDPG \citep{Lillicrap2016ContinuousCW} policies in DMControl. We consider the same tasks as PVR, but average Adroit results over two camera views to improve robustness of results (PVR only considers two tasks in this domain). Results for individual views are shown in Figure \ref{fig:appendix-adroit-cameras}. Our real robot demonstrations are collected via human teleoperation. PVR did not conduct real-world experiments. \emph{Our key difference compared to the original LfS baseline in PVR is the use of random shift augmentation}; all other implementation details remain identical to the original paper.

\subsection{On-Policy RL}
\label{sec:appendix-onpolicy-rl}
We closely follow the implementation of MVP \citep{Xiao2022MaskedVP}. Observations are single $224\times224$ RGB images and also include proprioceptive state information. The original LfS baseline proposed in MVP uses a small ViT \citep{Dosovitskiy2021AnII} encoder. We propose \emph{two} improved LfS baselines for this setting: \emph{(1)} {an LfS baseline that uses a shallow ConvNet encoder and \emph{no} data augmentation}, referred to as \emph{LfS}, and \emph{(2)} {an LfS baseline that additionally applies random shift augmentation} to the input images, referred to as \emph{LfS (+aug)}. Our proposed LfS encoder for on-policy RL can be summarized as:
{\small
\begin{codesnippet}
(0): Conv2d(3, 32, kernel_size=(7, 7), stride=2)
(1): ReLU()
(2): Conv2d(32, 32, kernel_size=(5, 5), stride=2)
(3): ReLU()
(4): Conv2d(32, 32, kernel_size=(3, 3), stride=2)
(5): ReLU()
(6): Conv2d(32, 32, kernel_size=(3, 3), stride=2)
(7): ReLU()
(8): Conv2d(32, 32, kernel_size=(3, 3), stride=2)
(9): ReLU()
(10): Conv2d(32, 32, kernel_size=(3, 3), stride=2)
(11): ReLU()

\end{codesnippet}
}
Following the MVP implementation, output feature maps are flattened and passed through a LayerNorm and linear projection:
{\small
\begin{codesnippet}
x = encoder(x)
x = x.view(x.shape[0], -1)
x = Linear(LayerNorm(x))

\end{codesnippet}
}

Meanwhile, following previous work on visual RL \citep{Hansen2021StabilizingDQ, yarats2021mastering}, we add an additional trunk layer to the policy head:
{\small
\begin{codesnippet}
(0): Linear(Z, Z)
(1): nn.LayerNorm(Z)
(2): nn.Tanh()

\end{codesnippet}
}
where \texttt{Z} denotes the dimensionality of image features concatenated with the proprioceptive state. We apply random shift augmentation to inputs using a padding of 10 to keep the padding-to-image ratio consistent with its original proposal. Following prior work \citep{Hansen2021StabilizingDQ, raileanu2020automatic}, we do not augment value targets. All other implementation details are kept identical. As our results in Figure \ref{fig:main-result} reveal, data augmentation is not necessary for on-policy RL algorithms such as PPO, and both of our two LfS baselines thus improve over the original baseline.

\subsection{Off-Policy RL}
\label{sec:appendix-offpolicy-rl}
We closely follow the implementation of DrQ-v2 \citep{yarats2021mastering} -- a state-of-the-art LfS method that uses random shift augmentation -- for our off-policy RL experiments. Observations are stacks of $84\times84$ RGB images ($3$ frames) with no access to state information. We denote this baseline as \emph{LfS (+aug)} since it already employs augmentation by default, and construct our \emph{LfS} baseline by simply disabling augmentation in DrQ-v2. We make no changes to the architecture nor hyperparameters, but list the encoder here for completeness:
{\small
\begin{codesnippet}
(0): Conv2d(9, 32, kernel_size=(3, 3), stride=2)
(1): ReLU()
(2): Conv2d(32, 32, kernel_size=(3, 3), stride=1)
(3): ReLU()
(4): Conv2d(32, 32, kernel_size=(3, 3), stride=1)
(5): ReLU()
(6): Conv2d(32, 32, kernel_size=(3, 3), stride=1)
(7): ReLU()

\end{codesnippet}
}

\section{Data Augmentation}
\label{sec:appendix-data-augmentation}
\begin{figure}[t]
    \centering
    \begin{subfigure}[b]{0.48\textwidth}
        \centering
        \includegraphics[width=\textwidth]{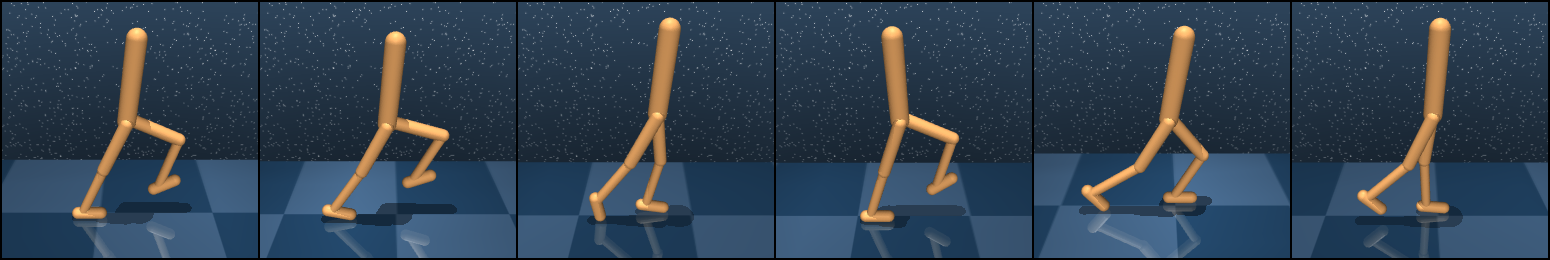}
        \caption{No augmentation.}
        \vspace{0.1in}
    \end{subfigure}
    \begin{subfigure}[b]{0.48\textwidth}
        \centering
        \includegraphics[width=\textwidth]{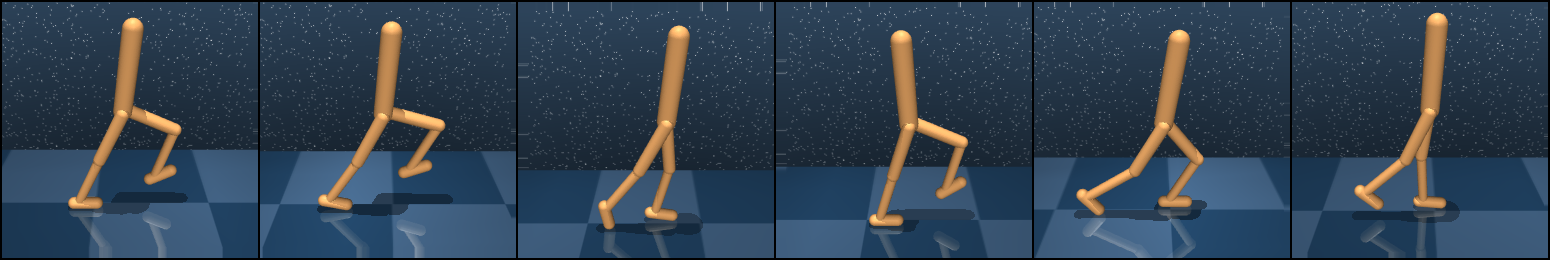}
        \caption{Random shift.}
        \vspace{0.1in}
    \end{subfigure}
    \begin{subfigure}[b]{0.48\textwidth}
        \centering
        \includegraphics[width=\textwidth]{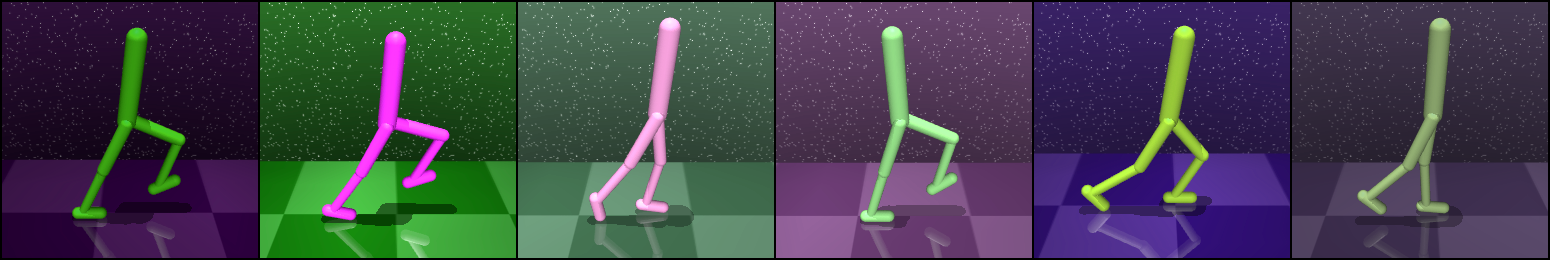}
        \caption{Color jitter.}
    \end{subfigure}
    \begin{subfigure}[b]{0.48\textwidth}
        \centering
        \includegraphics[width=\textwidth]{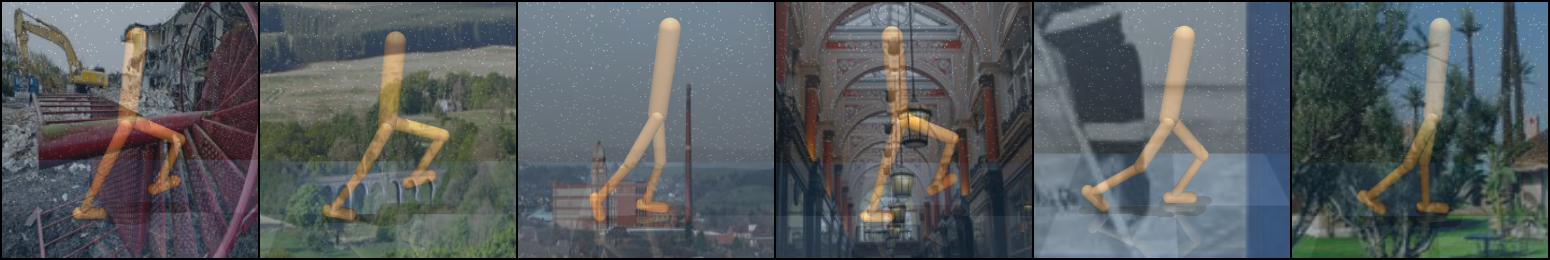}
        \caption{Random overlay.}
    \end{subfigure}
    \vspace{-0.125in}
    \caption{\textbf{Data augmentation}. Visualization of all choices of data augmentation considered in this work. We adopt augmentation hyperparameters from prior work without modification.}
    \label{fig:appendix-data-augmentation}
\end{figure}

We consider three choices of data augmentation in this work: random \emph{image shift} \citep{Kostrikov2021ImageAI, yarats2021mastering}, random \emph{color jitter}, and random \emph{overlay} \citep{hansen2021softda}. Augmentations are visualized in Figure \ref{fig:appendix-data-augmentation}. As in prior work, augmentations are applied consistently across time when using frame stacking. For completeness, we also visualize sample environments from the two test domains from DMControl Generalization Benchmark \citep{hansen2021softda} in Figure \ref{fig:appendix-dmcgb}, for which a sizable domain gap remains even after applying data augmentation to observations. In our RL experiments on visual robustness (strong augmentation), we use the objective of \citet{Hansen2021StabilizingDQ} to stabilize training.

\end{document}